**How Do Pedestrians' Perception Change toward Autonomous Vehicles during Unmarked Midblock Multilane Crossings: Role of AV Operation and Signal Indication**

**Fengjiao Zou, Ph.D. (Corresponding author)**
Transportation Engineer
Arcadis U.S., Inc.
Email: fengjiz@clemson.edu

**Jennifer Harper Ogle, Ph.D.**
Professor and Department Chair
Glenn Department of Civil Engineering, Clemson University
Email: ogle@clemson.edu

**Patrick Gerard, Ph.D.**
Professor
Department of Mathematical Sciences, Clemson University
Email: pgerard@clemson.edu

**Weimin Jin, Ph.D.**
Traffic and ITS Engineer
Arcadis U.S., Inc.
Email: Weimin.Jin@arcadis.com

**ABSTRACT**
One of the primary impediments hindering the widespread acceptance of autonomous vehicles (AVs) among pedestrians is their limited comprehension of AVs. This study employs virtual reality (VR) to provide pedestrians with an immersive environment for engaging with and comprehending AVs during unmarked midblock multilane crossings. Diverse AV driving behaviors were modeled to exhibit negotiation behavior with a yellow signal indication or non-yielding behavior with a blue signal indication. This paper aims to investigate the impact of various factors, such as AV behavior and signaling, pedestrian past behavior, etc., on pedestrians' perception change of AVs. Before and after the VR experiment, participants completed surveys assessing their perception of AVs, focusing on two main aspects: "Attitude" and "System Effectiveness." The Wilcoxon signed-rank test results demonstrated that both pedestrians' overall attitude score toward AVs and trust in the effectiveness of AV systems significantly increased following the VR experiment. Notably, individuals who exhibited a greater trust in the yellow signals were more inclined to display a higher attitude score toward AVs and to augment their trust in the effectiveness of AV systems. This indicates that the design of the yellow signal instills pedestrians with greater confidence in their interactions with AVs. Further, pedestrians who exhibit more aggressive crossing behavior are less likely to change their perception towards AVs as compared to those pedestrians with more positive crossing behaviors. It is concluded that integrating this paper's devised AV behavior and signaling within an immersive VR setting facilitated pedestrian engagement with AVs, thereby changing their perception of AVs.

# INTRODUCTION

Autonomous vehicles (AVs) can reduce driver involvement starting from level 3 of conditional driving automation, and at level 5 of full driving automation, they can eliminate driver involvement altogether, as per (1). According to the National Highway Traffic Safety Administration (NHTSA), 94% of motor vehicle crashes in the United States involved driver error (2). By removing drivers from the equation, AVs are expected to improve safety in the transportation system and society (3). However, a major challenge faced by manufacturers and organizations aiming to introduce AVs to the market is the lack of public understanding or trust in AVs and their associated technologies (4). As yet, most of the attention on human perception of AVs has focused on the perspective of drivers (4,5). Meanwhile, pedestrians, as the most vulnerable road users, have received less attention regarding their perception of AVs. The extant literature on how the interaction between AVs and pedestrians may influence the latter's perception of AVs is exceedingly scarce. As most pedestrians have not interacted with AVs and have difficulty envisioning such interactions, it is crucial to provide them with an environment that closely resembles reality in order to better understand AVs and, possibly, change their perception of AVs. Further, by understanding how interactions with AVs influence pedestrian attitudes, developers and policymakers can address concerns and design strategies to increase AV acceptance and adoption rates.

Virtual reality (VR) has recently emerged as a valuable tool for studying pedestrian behavior, as demonstrated by several studies (6–11). However, only a few researchers have investigated how pedestrians behave in response to various AV operations and behaviors. This study employs VR to provide pedestrians with an immersive environment for engaging with and comprehending AVs. A challenging roadway type and location was selected for the experiment - an unmarked midblock area on a multilane road, where pedestrian crash rates have been shown to be high (12). Also, at the midblock unmarked road, there is concern that if AVs are programmed to always be conservative, pedestrians may take advantage of them, potentially influencing traffic systems and ultimately impacting public perception of AVs (13,14). Therefore, this study modeled diverse AV driving behaviors to exhibit negotiation behavior with a yellow signal indication or non-yielding behavior with a blue signal indication at unmarked midblock locations. Before and after the VR experiment, participants completed surveys assessing their perception of AVs, focusing on two main aspects: "Attitude" and "System Effectiveness." Other surveys like AV behavior and signaling, pedestrian demographics, pedestrian past behavior, pedestrian walking exposure, VR sickness, and VR presence were also provided to all participants. The aim is to investigate the impact of the various factors on pedestrians' perception change of AVs. Questions of interest are, "Which pedestrians are more likely to change their perceptions? And what are their prior stated experiences as a pedestrian?".

**Contributions**
This study has unique aspects that add to the existing research. These include 1) the use of VR to investigate pedestrians' perception changes of AVs, 2) the examination of the impact of factors, like AV behavior and signaling, pedestrian past behavior, etc., on these perception changes, and 3) the involvement of designing various AV operations, including negotiation behavior and non-yielding behavior; and designing AV adapting behavior based on real-time analysis of pedestrians' movements.

**LITERATURE REVIEW**
Although several studies have been conducted to examine the perception of AVs or automation, a significant portion of these investigations has overlooked the pedestrians' standpoint. Nonetheless, these studies provided valuable insights into the factors that influence the perception of AVs. According to Schaefer et al., individual characteristics are the most significant factor in building trust toward automation, which implies that even a well-designed automated system may not necessarily gain humans' trust (15). Some studies have focused on the perception of AVs from the driver's perspective. Khastgir et al., for instance, conducted a study using a driving simulator with 56 participants and found that drivers increased their trust in the automated system after learning about its strengths and limitations (16).

**Pedestrian Perception of AV**
Only a few studies have examined the perception of AVs from the perspective of pedestrians. Deb et al. developed a questionnaire to measure pedestrian receptivity to AVs based on attitude, social norms, trust, compatibility, and system effectiveness (17). It was found that male and younger respondents, those from urban areas, and those with more positive pedestrian behavior had higher receptivity to AVs. Although some researchers were concerned that pedestrians could find it difficult to imagine the specific traffic condition and interact with AVs (18), the surveys Deb et al. designed are still a vital source of inspiration for pedestrian perception of AV studies. Another study conducted by Reig et al. was one of the first to observe pedestrians' interaction with AVs operating in the real-world (19). The authors interviewed 32 pedestrians in the US who had interacted with AVs. They found that most of those participants had heard little about AVs before, and the less knowledge participants had of AVs, the lower scores they rated for perceived importance, interest, and trust of AVs. However, opportunities for pedestrians to interact with real-world AV testing are limited, and it is challenging to test different AV behaviors and collect pedestrian perceptions.

Recently, VR has gained popularity in human behavior studies (7), and some researchers have used it to study pedestrian perception of AVs (8,18). Many studies have compared the effects of external human-machine interface (eHMI) versus no interface. Böckle et al. and Mahadevan et al. found that using eHMI significantly led to a more comfortable interaction between pedestrians and AVs (20,21). Other researchers came to the conclusion that pedestrians view AVs with eHMI to be safer compared to those without it (9,20,22,23). Some studies did not directly investigate pedestrians' perception of AVs; they explored pedestrian behavior when interacting with AVs, which also reflected their perception, e.g., waiting time and crossing time. It was found that eHMI significantly decreased pedestrian waiting times and allowed for quicker crossing decisions (6,10,11,23–27). Some studies examined pedestrian perception of AVs with different driving behavior (28) and found that aggressive driving behavior of AVs negatively impacted pedestrians' trust, particularly at unsignalized crosswalks. Overall, the investigation of pedestrian perception concerning AVs remains a relatively nascent subject when compared to the more established general public perspective of AVs. The extant literature on how the interaction between AVs and pedestrians may influence the latter's perception of AVs is exceedingly scarce.

**AV Operation and Signal Design**
In order to study pedestrians' perception of AVs in VR, it is essential to have a well-designed AV operation system. Previous studies have mainly focused on different types of eHMI or signals, such as text messages, smiley faces, traffic lights, or a walking man (22,25,27), but few have taken AV behavior into account. Some experiments have only programmed AVs to operate at a specific velocity or to slow down for pedestrians at crosswalks based on a pre-established rate of deceleration (23,29), while others have programmed AVs to never give way to pedestrians (6). Only a small number of researchers have investigated how pedestrians behave in response to various AV operations. Some demonstrated how an early deceleration could enhance the comfort level of pedestrians (26,30), increase pedestrian trust in the AV system (28), and improve traffic efficiency because AVs could accelerate again without having to make a complete stop (26). However, those studies did not permit AV with real-time and reciprocal communication with pedestrians. Only one study, conducted by (14), continuously adjusted the AV behavior to align with the participants' movements, which is considered a more realistic approach to AV behavior in the future, according to (8). Camara et al. used VR to evaluate the preferences of pedestrians for AV driving styles, specifically game theoretic driving (14).

**Research Gap**
In brief, the literature suggests that there has been limited research on pedestrians' perception of AVs. Additionally, while some studies have explored changes in drivers' perceptions of AVs, few have looked at pedestrians' perception changes, as well as the factors that affect these changes. For AV operation and signaling design, only a few researchers have investigated how pedestrians behave in response to various AV operations. Besides, only one study has continuously adapted AV behavior based on pedestrians' movements.

**METHODOLOGY**
The methodology includes the VR experiment design, survey design, data collection procedure, and statistical models.

**VR Experiment Design**
The design of the VR experiment involved modeling AV behavior and signaling and designing pedestrian tasks. The VR environment was developed using Unity 3D software, and participants used an Oculus Quest 2 headset during the experiment. More experimental design details can be found in (10).

*AV Behavior and Signals Designs*
As mentioned in the introduction, pedestrians are not always given priority when crossing the road, particularly at midblock locations without marked crosswalks. Some researchers have suggested that future AVs need to negotiate the right of way with pedestrians (14,31). In this study, AVs were modeled that do not always yield to pedestrians at unmarked midblock locations. Two different AV behaviors and signals were designed: yellow signals indicating real-time negotiation behavior that will yield under certain circumstances and blue signals indicating a platoon of AVs that will not yield to pedestrians. A baseline scenario (No Signal) in which AVs operate the same as those with yellow signals was also included; however, they won't show any signal indications. In the absence of communication cues, pedestrians must rely on implicit AV behavior, such as vehicle speed and distance, to make crossing decisions. AVs were modeled with a speed limit of 20 mph, and no turning maneuvers were included. Furthermore, all AV behavior and signal scenarios were randomized with different types of roadways during the experiment.

1. Yellow Signal Design

In this study, AVs were modeled with negotiation behavior that displays a yellow signal. When an AV detects the presence of a pedestrian and identifies a trajectory conflict with the AV's path, the AV initiates a negotiation process to establish the right of way with the pedestrian, indicated by a yellow signal. The AV will yield only if the pedestrian reaches the conflict box area (a rectangular box with the same width as the AV along the pedestrian's travel path). However, if the pedestrian has not reached the conflict box area or hesitates and remains stationary, the AV wins the right of way and continues at its current speed. The decision-making process of AVs is contingent upon the response of pedestrians, who determine whether to yield or not yield at each timestamp, thereby influencing the AV's subsequent actions. As shown in **Figure 2**, the yellow signal was placed in front of the AV to communicate its intention to pedestrians. More details of the yellow signal design and AV behavior algorithm can be found in (10).

2. Blue Signal Design

Platoons of AVs with blue signals that do not yield to pedestrians at unmarked midblock locations were also designed. A closely operating platoon of AVs with the same speed can benefit the traffic system at unmarked midblock locations. So, at such locations, the platoon of AVs will take priority ongoing and will not yield to the pedestrian. An example of the platoon of AVs with blue signals mixed with AVs with yellow signals is shown in **Figure** 1. Following the platoon of blue AVs, there will be a few AVs with yellow signals coming that negotiate with pedestrians, and another platoon of AVs with blue signals follows behind.

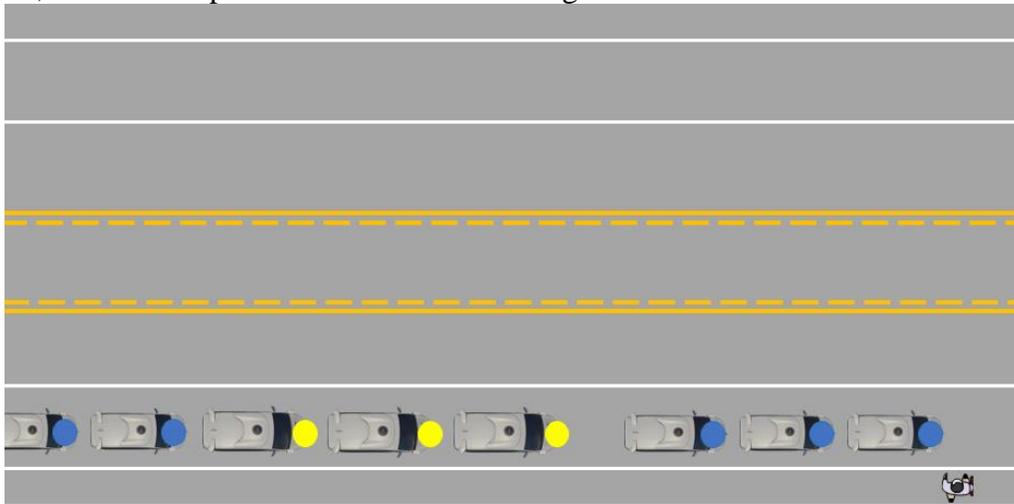

**Figure 1 Representative platoon of AVs with blue signal mixed with AVs with yellow signal**

*Pedestrian Task*
**Figure 2** displays an instance of a participant waiting to cross a four-lane road with a central turning lane in both the VR environment and real life. In this example, the participant faces an AV displaying a yellow signal, indicating negotiation between the AV and the pedestrian. The participants were instructed to pick up food from a restaurant across the multilane road and return home. The closest crosswalk was far away, so they decided to jaywalk. Participants were told that all vehicles operating on the road were AVs. To enable participants to stay in the VR environment longer without being interrupted by researchers, randomized scenarios were created that would be generated automatically for the participants after each task completion. Further details regarding the design of the road can be found in (10).

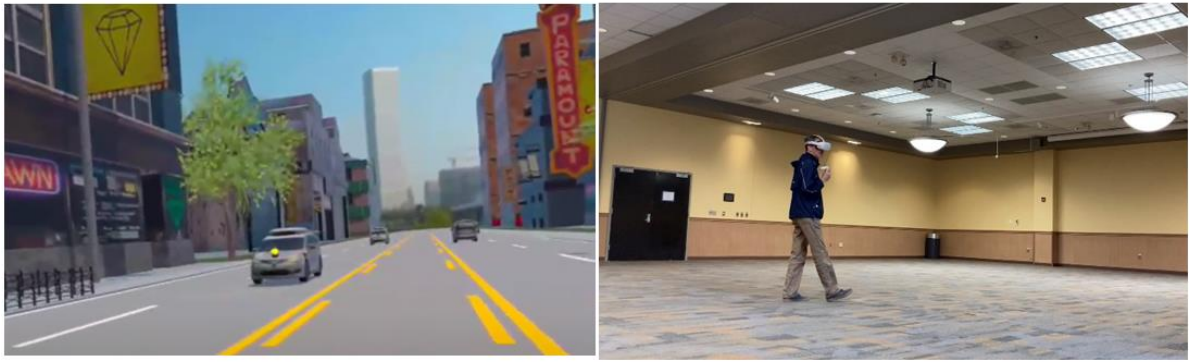

(a): Pedestrian waiting in VR          (b): Pedestrian waiting in reality

**Figure 2 Pedestrian in VR (a) and reality (b)**

**Surveys Design**

During the experiment, participants were asked to complete several surveys to study factors affecting their perception changes of AVs.

*Perception of AVs*

A set of eight questions (listed in **Figure 3**) was developed to assess pedestrians' perceptions of AVs. Questions 1-7 were adapted from a previous study (17), and question 8 was added to address concerns raised by other researchers about pedestrian occlusion issues (32,33). Participants were asked to rate their responses on a five-point scale ranging from 1 (strongly disagree) to 5 (strongly agree). Questions 1-4 aimed to assess pedestrians' overall attitude when interacting with AVs while crossing the road, while questions 5-8 assessed the effectiveness of AVs in detecting and interacting with pedestrians. The survey questions were administered to participants both before and after the VR experiment to gauge any changes in their perceptions.

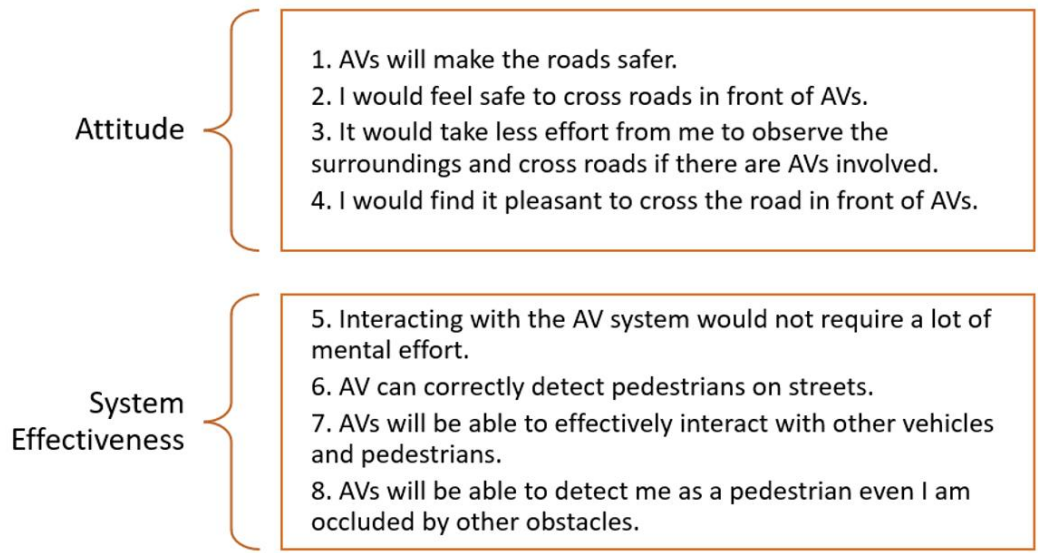

**Figure 3 Perception of AVs survey items**

*AV Behavior and Signaling*
The behavior and signaling of AVs are crucial factors affecting pedestrians' perception of AVs in this study. As outlined in the VR experiment design, two main types of AV behavior and signaling were designed: a yellow signal indicating an AV with negotiation behavior and a blue signal indicating a platoon of non-stopping AVs at the midblock location. Four questions related to AV behavior and signaling were posed to participants after they completed the VR experiment, shown in **Figure 4**. Participants rated their responses on a five-point scale.

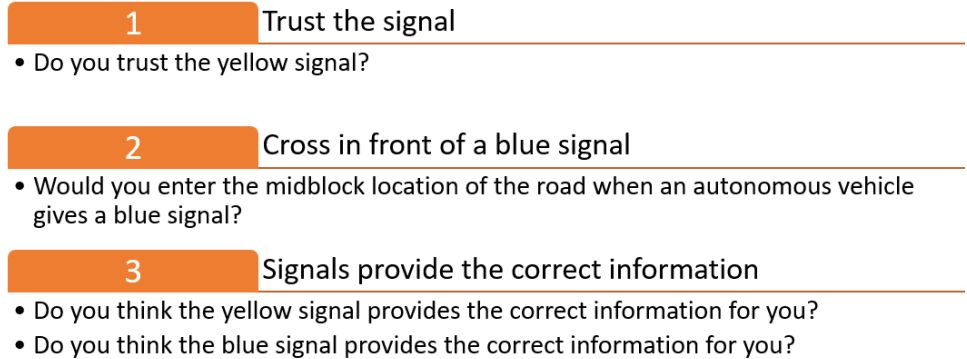

**Figure 4 AV behavior and signaling survey items**

*Pedestrian Past Behavior*
The 20-item general pedestrian past behavior scale (17) was utilized to categorize participants' behaviors as pedestrians. The pedestrian past behavior comprises five subscales, which represent five categories of pedestrian behavior: violation, error, lapse, aggressive behavior, and positive behavior. The pedestrian past behavior was designed to determine if pedestrian behavior affects their perception of AVs. The questionnaire includes items such as "I cross outside the pedestrian crossing even if there is one (crosswalk) less than 50 meters away", and "I cross between vehicles stopped on the roadway in traffic jams." Participants rated their responses on a five-point scale, ranging from 1 (never) to 5 (always)."

*Pedestrian Walking Exposure*
Pedestrian walking exposure was also collected, including the number of utilitarian walking trips per day, recreational walking trips per week, daily walking time, and whether pedestrians have the proper pedestrian infrastructures for where they walk.

*VR Sickness*
The simulator sickness questionnaire (SSQ), developed by (34), was employed to monitor participants' motion sickness throughout the study. The SSQ comprises sixteen symptoms, including general discomfort, fatigue, headache, eye strain, difficulty focusing, increased salivation, sweating, nausea, difficulty concentrating, fullness of head, blurred vision, dizziness (eyes open), dizziness (eyes closed), vertigo, stomach awareness, and burping. After completing the training and each experimental block, participants were asked to rate the severity of their symptoms on a scale from none (represented by 0) to severe (represented by 3).

*VR presence*
At the end of the experiment, a 15-item version of the presence questionnaire (version 2.0) was used to evaluate the level of immersion experienced by the participants in the VR environment (35). The 15 questions include 12 items (3, 5, 8, 9, 12-13, 18, and 23-27) directly adapted from (35). Item 14 was modified from (35) to "how completely you were able to obtain information about traffic in the environment using vision." Additionally, item 28 was modified to "how well the visual display quality assisted you in performing assigned tasks." Finally, a new item that focused on the clarity and legibility of images and text during the experiment was included.

**Data Collection Procedure**
**Figure 5** below shows the data collection procedure, where the parts related to the surveys are highlighted in bold. The participants first filled out surveys related to their demographics, pedestrian past behavior, pedestrian past walking exposure, and perception of AVs. Next, the participants received a training session in VR to understand the task, get familiar with the VR equipment, comprehend the meaning of yellow and blue signals, and interact with AVs displaying different behaviors and signals. The data collection only began after the participants fully understood the AV signals during the training sessions. The participants then completed three repeated experimental blocks, each with nine randomized trials with different AV signals and roadway scenes. Each participant completed a total of 27 trials. After the training and after each block, participants completed the 16-item SSQ to monitor their well-being. In the end, participants completed the perception of AV questions one more time and the AV behavior and signaling and VR presence questions. More details of the data collection procedure can be found in (10,11).

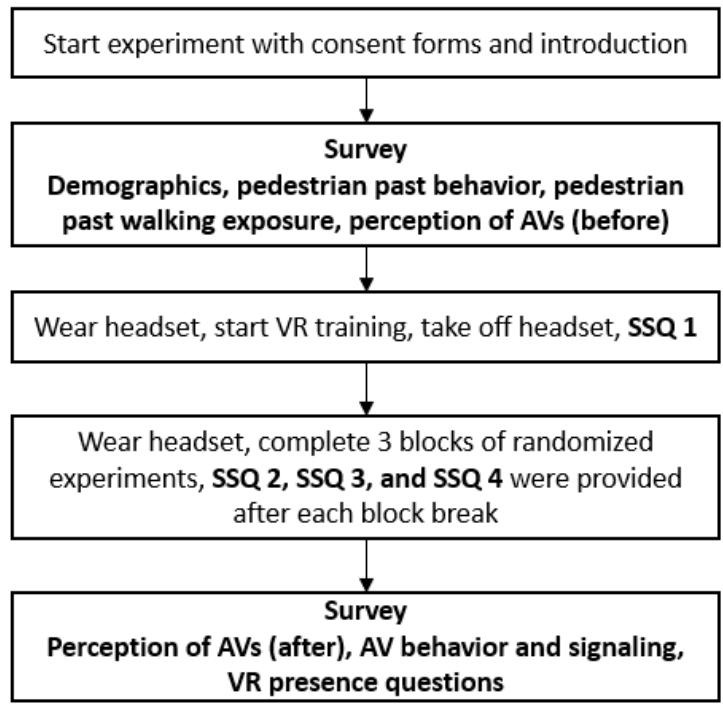

**Figure 5 An overview of the data collection procedure**

Fifty participants were recruited from Clemson University and the surrounding community, including 31 males and 19 females. The age range of the participants varied from 18 to over 60 years old, with 33 participants falling within the 18-29 age range, six participants within the 30-39 age range, six participants within the 40-49 age range, one participant within the 50-59 age range (ultimately combined with 40-49 age range), and four participants over the age of 60. No one dropped the experiment due to motion sickness. The overall average sickness score of 5.788 was considered minimal sickness symptoms based on (36). The average VR presence score of 4.33 out of 5 represents that participants were involved in the VR environment during the data collection process compared to other related studies (9). Each participant received a $20 gift card for their time.

**Statistical Models**
*Wilcoxon Signed-rank Test*
This paper applies the Wilcoxon signed-rank test, a non-parametric test, for the before and after perception changes analysis. Wilcoxon signed-rank test is usually used for the analysis of matched-pair data, and the data is not normal (37). In this study, the dependent variables (perception of AVs) were rated from 1 to 5 before and after, so the data is not normally distributed. The Null Hypothesis is that in the population, the central tendency between the paired data is zero (37).

*Linear Regression Model*
A simple linear regression model is used to explore the relationship between dependent variable perception changes and independent variables like demographics, pedestrian past behavior, pedestrian past walking exposure, AV behavior and signaling, and VR experience. The simple linear regression model is represented as follows (38),
$$y = X\beta + \varepsilon$$
Where, $y$ is a $N \times 1$ vector of the dependent variable. $X$ represents a $N \times p$ design matrix of $p$ explanatory variables. $N$ is the number of observations in the dataset used in the model. $\beta$ represents a $p \times 1$ vector of the regression coefficients; $\varepsilon$ is a $N \times 1$ vector of the residuals. R software is utilized to run the simple linear regression model using "lm" package (39).

**RESULTS**
The analysis begins by presenting descriptive statistics for survey questions related to the perception of AVs, both before and after the VR experiment. These questions were then categorized into two distinct aspects: "Attitude" and "System Effectiveness". Wilcoxon signed-rank tests were employed to determine whether there was a significant difference between the means of "Attitude" and "System Effectiveness" before and after the experiment. Finally, various factors were examined to understand their potential impact on pedestrian perception changes regarding AVs. These factors included AV behavior and signaling, demographics, pedestrian past behavior, walking exposures, and the participant's VR experience.

**Descriptive Statistics**
**Figure** 6 below depicts the changes in participant perception of AVs before and after the VR experiment. The figure demonstrates that, overall, there was a positive change in perception after participants interacted with AVs in the VR experiment. For example, participants were more likely to believe that AVs would make the roads safer, and they also felt safer and more comfortable crossing in front of AVs.

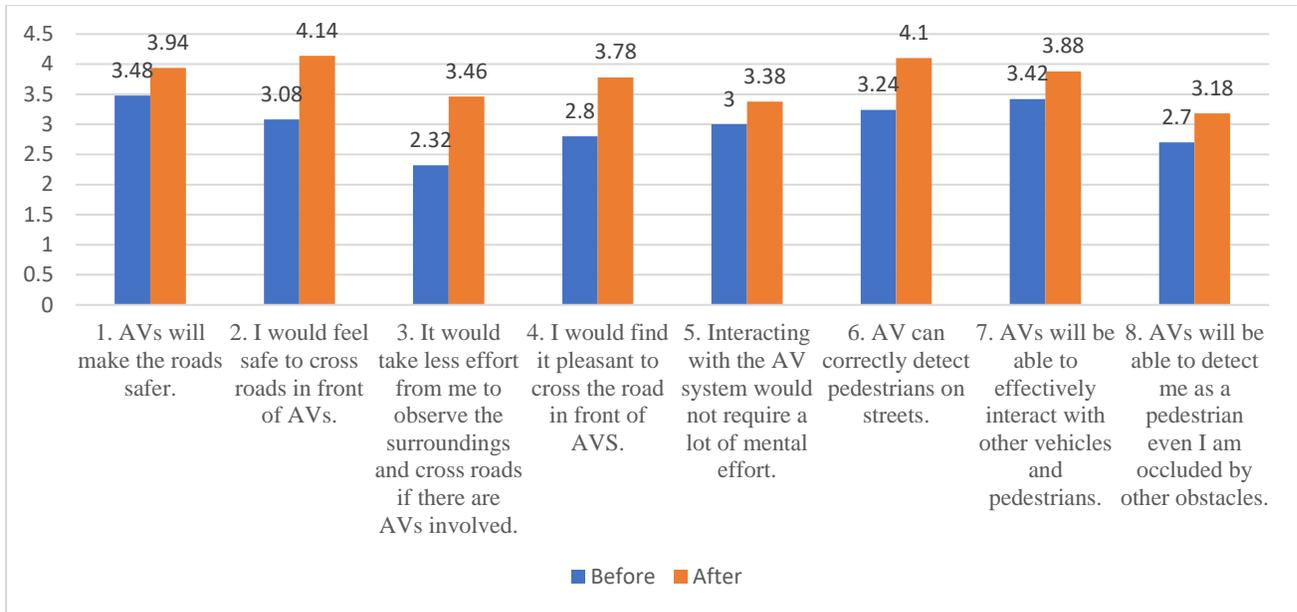

**Figure 6 Perception of AVs before and after the VR experiment**

**Perception of AVs (before-After)**
To test the perception changes before and after the VR experiment, questions 1 to 4 were grouped as "Attitude" toward AVs, which reflects participants' positive or negative feelings towards AVs in general. Questions 5 to 8 were combined as "System Effectiveness," which measures the extent to which participants trust that AVs can accurately detect and interact with pedestrians. In the rest of the paper, the perception of AVs will be based on those two aspects: "Attitude" and "System Effectiveness." Wilcoxon signed-rank test was applied, and the null hypothesis was that the central tendencies of the two dependent samples were the same in the population. The results, as shown in **Table 1**, reveal that the p-values for both "Attitude" and "System Effectiveness" change were less than 0.05, indicating that the null hypothesis was rejected and the before and after mean differences were not the same. **Table 1** further indicates that for both "Attitude" and "System Effectiveness", the means after the VR experiment were greater than the means before the experiment. The larger after-experiment mean values imply that pedestrians' overall scores of attitude toward AVs and trust in the effectiveness of the AV system were significantly increased after the VR experiment.

**Table 1 Perception of AVs (before-after)**

| Factors Evaluated | Mean | | Wilcoxon signed-rank test | |
|---|---|---|---|---|
| | Before | After | V value | p-value |
| Attitude Change | 2.92 | 3.8 | 883 | < 2.2e-16 |
| System Effectiveness Change | 3.09 | 3.635 | 2288.5 | 1.17e-08 |

Although **Table 1** indicates that overall scores of pedestrians' perception of AVs improved, including their general attitude and trust in the effectiveness of AV systems, it is noted that some participants gave lower scores on the survey questions after the VR experiment. For example, six out of fifty participants rated question 1, "AVs will make the roads safer," lower after the *Zou, Ogle, Gerard, Jin*experiment. For questions 2, 3, 4, and 6, fewer than five participants rated their scores lower. For questions 7 and 9, nearly ten participants rated their scores lower. Notably, for question 5, fifteen participants rated their scores lower after the experiment. To investigate the factors that may have influenced these differing perceptions, the analysis utilized simple linear regression models.

**Factors Affecting the Perception Changes Toward AVs**
This section presents all the factors tested using simple linear regression models to explore pedestrians' perception changes toward AVs. **Table 2** presents the findings of the impact of AV behavior and signaling, pedestrian past behavior, and VR experience on pedestrians' perception changes toward AVs.

*AV Behavior and Signaling*
The results in **Table 2** reveal that trust in the yellow signal has a significant effect on pedestrians' perception changes toward AVs, including their overall attitude and trust in the system's effectiveness. For attitude, the estimate of 1.438 indicates that the more pedestrians trust the yellow signal, the more likely they are to increase their general attitude score towards AVs following the VR experiment. Similarly, for system effectiveness, pedestrians who trust the yellow signals are more likely to increase their trust in the AV system's effectiveness after the VR experiment. Additionally, the results of question 2 indicate that entering the midblock location of the road when an AV gives a blue signal significantly affects pedestrians' attitude towards AVs at a 0.1 significance level. The estimate of -0.773 suggests that the more pedestrians intend to enter the midblock location when AV gives a blue signal, meaning non-stopping behavior, the less likely they are to change their attitude toward AVs after the VR experiment. Question 3 produced similar results to question 1, with the yellow signal providing the correct information being significant in the model at a 0.01 significance level. Those who believed the yellow signal provided the correct information during the VR experiment were more likely to have a positive attitude towards AVs and trust the AV system's effectiveness more.

**Table 2 Effects of Factors that Affect Perception Changes Toward AVs**

|  | Attitude Change |  | System Effectiveness Change |  |
|---|---|---|---|---|
|  | Mean/ estimates | p-value | Mean/ estimates | p-value |
| **AV Behavior and Signaling** | | | | |
| 1. Trust yellow signal | 1.438 | 0.011* | 1.612 | 0.005** |
| 2. Enter midblock with blue signal | -0.773 | 0.073. | -0.197 | 0.657 |
| 3. Yellow signal provides the correct information | 1.596 | 0.005** | 1.485 | 0.010** |
| 4. Blue signal provides the correct information | 0.676 | 0.157 | 0.702 | 0.146 |
| **Pedestrian Past Behavior** | | | | |
| Violation | -0.600 | 0.292 | -0.076 | 0.896 |
| Error | -0.940 | 0.305 | -0.761 | 0.413 |
| Lapse | -0.990 | 0.381 | -0.313 | 0.785 |
| Aggressive behavior | -0.209 | 0.859 | 0.299 | 0.802 |
| Positive behavior | 1.911 | 0.013* | 1.646 | 0.035 * |

| **VR Experience** | | | | |
|---|---|---|---|---|
| VR motion sickness | -0.046 | 0.174 | -0.067 | 0.049* |
| VR presence | -0.071 | 0.946 | -0.566 | 0.590 |

Results with . are significant at a 0.1 level
Results with * are significant at a 0.05 level
Results with * *are significant at a 0.01 level

*Pedestrian Past Behavior*
**Table 2** also shows the effects of pedestrian past behavior, including violation, error, lapse, aggressive behavior, and positive behavior. Results show that for both "Attitude" and "System Effectiveness", the only significant variable is "positive behavior" at a 0.05 significance level. Pedestrian violation, error, lapse, and aggressive behavior were not significant in the model. Taking "Attitude" as an example, the estimate of 1.911 for positive behavior means that for one unit change in the positive behavior variable, the "Attitude" change will be increased by 1.911. This result suggests that pedestrians who show positive behavior are those who are more likely to increase their general attitude score toward AVs after the VR experiment. Similarly, for System effectiveness, pedestrians who show positive behavior are more likely to increase their trust in the AV system's effectiveness after interacting with AVs in the VR experiment.

*VR Experience*
In order to understand how participants' experiences with VR impact their perception of AVs, data were collected on both VR motion sickness and VR presence scores. For VR motion sickness, the SSQ score after the final experiment block was used to measure participants' overall sickness in the VR experiment. Results in **Table 2** show that motion sickness does not affect pedestrians' overall attitude score change toward AV. However, the degree of motion sickness experienced by participants did significantly affect their trust in the AV system's effectiveness. The mean of -0.067 indicates that the more motion sickness the participants got in the VR experiment, the less change of trust they had in the AV system's effectiveness. Overall, results suggest that experiencing motion sickness during a VR experiment did not necessarily result in a negative attitude toward AVs among pedestrians. However, it did impact their trust in the AV system's effectiveness. This finding highlights the importance of carefully designing VR experiments to avoid biased results due to participants' motion sickness. Additionally, results show that participants' sense of presence in the VR experiment did not significantly impact their perception of AVs.

*Other tests with insignificant findings*
Other factors were also tested in the model. Results show that both age and gender were not significant in the model. Pedestrian walking exposure was also tested. Results show that none of the past walking exposure factors are significant in the models, meaning that pedestrians' walking exposure does not affect how they change their perception of AVs. The results imply that no matter how much pedestrians walk in the past and where they walk, the interaction with AV is still a new concept and experience to them. So, pedestrians' overall perception of AVs depends more on their experience with AVs than their past walking exposure.

**CONCLUSION AND DISCUSSION**
This paper aims to address one of the main challenges to the popularity of AVs related to pedestrians, which is pedestrians' lack of trust and understanding of AVs and related technologies. To achieve this, the study used VR to provide pedestrians with an immersive environment to interact with and understand AVs. Specifically, an unmarked midblock location on a multilane road was designed in the VR environment, with AVs modeled to exhibit negotiation behavior or non-yielding behavior. Surveys regarding the perception of AVs were administered to pedestrians both before and after the VR experiment. Pedestrians were also provided with other surveys, including AV behavior and signaling, pedestrian demographics, pedestrian past behavior, pedestrian walking exposure, VR sickness, and VR presence. The aim is to investigate how those factors change pedestrians' perception of AVs.

To investigate the impact of the interaction with different AV behavior and signaling in VR on pedestrian perceptions of AVs, the study aggregated the AV perception survey questions into two aspects: "Attitude" and "System Effectiveness". The Wilcoxon signed-rank test was then used to compare the pre-and post-experiment perceptions. The results demonstrated that both the overall attitude score towards AVs and trust score in the effectiveness of AV systems were significantly increased after the VR experiment. These findings suggest that the designed AV system in an immersive VR environment allowed pedestrians to interact with AVs, which in turn increased their confidence in AV technology.

Simple linear regression models were employed to examine the factors that impact the changes in pedestrians' perception of AVs. The study first investigated AV behavior and signaling-related questions. The results revealed that the more pedestrians trust the yellow signals or perceive that the yellow signal provides correct information, the more likely they are to have a positive attitude towards AVs and to increase their trust in the AV system's effectiveness after the VR experiment. This finding implies that the design of the yellow signal, which indicates that AVs detect pedestrians and will negotiate and stop for them under certain circumstances, gives pedestrians more confidence and makes them feel more comfortable while interacting with AVs. Additionally, it was observed that pedestrians who are more likely to cross the road aggressively when an AV displays a blue signal (non-yielding) are less likely to change their general attitude towards AVs after the VR experiment. This finding suggests that pedestrians who exhibit more aggressive crossing behavior are less likely to change their perception towards AVs as compared to those pedestrians who display more positive crossing behaviors. Consequently, in endeavors aimed at mitigating concerns and devising effective strategies to enhance AV acceptance and adoption rates, developers and policymakers should tailor their approaches based on pedestrians' distinct characteristics.

This paper also explored pedestrian factors that affect pedestrian perception changes toward AVs, including pedestrian demographics, past behavior, and walking exposure. Results show that age and gender do not affect how pedestrians change their perception toward AVs. Pedestrian past behavior was also explored, including violation, error, lapse, aggressive behavior, and positive behavior. It's found that pedestrians who adhere to traffic regulations and display positive behavior toward other road users are more likely to exhibit a positive attitude toward AVs and increase their trust in the AV system's effectiveness after the VR experiment. This finding indicates that rule followers comprehend the design of AV signals and comply with the signal design rules, resulting in a better understanding and trust in AVs after the VR experiment. Pedestrian walking exposure was also tested, and results found that pedestrians' walking exposure does not affect how they change their perception of AVs. The results imply that no matter how much pedestrians walk in the past and where they walk, the interaction with AV is still a new concept and experience to them. So, pedestrians' overall perception of AVs is more dependent on their experience with AVs than their past walking exposure.

The effects of VR sickness on pedestrians' perception changes toward AVs were also investigated. It was discovered that if the experiment made pedestrians feel sick, they were less likely to have more trust in the AV system's effectiveness in detecting and interacting with pedestrians. This finding highlights the importance of carefully designing VR experiments to study pedestrians' trust, as motion sickness may lead to biased results.

It is concluded that the integration of this paper's specifically devised AV behavior and signaling within an immersive VR setting facilitated pedestrian engagement with AVs during unmarked midblock multilane crossings, thereby changing their perception of AVs. The participants' attitudes toward AVs and their trust in the efficacy of AV systems exhibited noteworthy increments subsequent to their participation in the VR experiment.

## ACKNOWLEDGMENTS

This work was partially supported by the Center for Connected Multimodal Mobility(C2M2) (the U.S. Department of Transportation Tier 1 University Transportation Center) headquartered at Clemson University, Clemson, SC, USA. Any Opinions, findings, conclusion, and recommendations expressed in this material are those of the author(s) and do not necessarily reflect the views of C2M2, and the US Government assumes no liability for the contents and use thereof.

## AUTHOR CONTRIBUTIONS

The authors confirm their contribution to the paper as follows: study conception: F. Zou, J. Ogle, P. Gerard, W. Jin; VR experiment design and data collection: F. Zou, J. Ogle, P. Gerard; analysis and interpretation of results: F. Zou, J. Ogle, P. Gerard, W. Jin; draft manuscript preparation: F. Zou, J. Ogle, W. Jin, P. Gerard. All authors reviewed the results and approved the final version of the manuscript.